%% file: root.tex
\documentclass[letterpaper, 10 pt, conference]{ieeeconf}  

\IEEEoverridecommandlockouts                              

\usepackage{cite}

\usepackage{amsmath}

\usepackage{siunitx}
\usepackage{diagbox}
\usepackage{booktabs}
\usepackage{graphics} 
\usepackage{epsfig} 
\usepackage{amsmath} 
\usepackage{amssymb}  
\usepackage[usenames,dvipsnames,svgnames,table]{xcolor} 
\usepackage{bm}
\usepackage[percent]{overpic}
\usepackage{lipsum}
\usepackage[font=footnotesize]{subfig}
\usepackage{graphicx}
\usepackage{blindtext}
\usepackage{caption}
\usepackage[normalem]{ulem}
\usepackage{url}
\usepackage{multirow}
\usepackage{hyperref}
\usepackage[switch]{lineno}
\usepackage{footmisc}

\newcolumntype{P}[1]{>{\centering\arraybackslash}p{#1}}

\usepackage{textcomp}
\usepackage{tikz}
\usetikzlibrary{shapes,arrows}
\usetikzlibrary{calc}
\usetikzlibrary{positioning}
\tikzset{%
  block/.style    = {draw, thick, rectangle, minimum height = 3em,
    minimum width = 3em},
  input/.style    = {coordinate}, 
  output/.style   = {coordinate} 
}

\input{math_commands}

\title{Convex strategies for trajectory optimisation: application to the Polytope Traversal Problem}

\author{Steve Tonneau
 \thanks{Steve Tonneau is at the University of Edinburgh, Scotland.}%
}

\begin{document}

\maketitle
\thispagestyle{empty}
\pagestyle{empty}

\begin{abstract}
Non-linear trajectory optimisation methods require good initial guesses to converge to a locally optimal solution. A feasible guess can often be obtained by allocating a large amount of time for the trajectory to be complete. However for unstable dynamical systems such as humanoid robots, this quasi-static assumption does not always hold.

We propose a conservative formulation of the trajectory problem that simultaneously computes a feasible path and its time allocation. The problem is solved as a convex optimisation problem guaranteed to converge to a feasible local optimum. 

The  approach is evaluated with the computation of feasible trajectories that traverse sequentially a sequence of polytopes. We demonstrate that on instances of the problem where quasi static solutions are not admissible, our approach is able to find a feasible solution with a success rate above $80 \%$ in all the scenarios considered, in less than 10ms for problems involving traversing less than 5 polytopes and less than 1s for problems involving 20 polytopes, thus demonstrating its ability to reliably provide initial guesses to advanced non linear solvers.

\end{abstract}

\input{body}


\bibliography{biblio}
\bibliographystyle{IEEEtran}
\end{document}

%% file: math_commands.tex
\newcommand{\vc}[1]{\mathbf{#1}} 					
\newcommand{\tbf}[1]{\textbf{#1}} 					
\newcommand{\sref}[1]{(Section \ref{#1})} 					
\newcommand{\fg}[1]{(Fig. \ref{#1})} 					
\newcommand{\fgg}[1]{Fig. \ref{#1}} 					
\newcommand{\eqq}[1]{(\ref{#1})} 					
\newcommand{\xv}[0]{\mathbf{x}} 					
\newcommand{\dex}[0]{\dot{\mathbf{x}}} 					
\newcommand{\ddx}[0]{\ddot{\mathbf{x}}} 					
 
\DeclareMathOperator*{\st}{\mathbf{s.t.}\,}					
\newcommand{\Expect}{{\rm I\kern-.3em E}}				



%% file: body.tex

\section{Introduction}

Trajectory generation is the computation of both a continuous path and its
time parametrisation, subject to geometric, continuity and dynamic constraints.
The problem is of importance in robotics, with applications in various fields including autonomous vehicles, manipulators,
UAVs and legged robots.

Trajectory generation is commonly addressed as a Trajectory Optimisation (TO) problem aiming to find the minimum time trajectory satisfying the constraints.
The non-linearity of the problem makes it hard to solve globally and efficiently, especially as real-time computation is often a requirement.

We consider the Polytope Traversal problem (PT), a TO problem where the trajectory is constrained to traverse a sequence of polytopes \fg{fig:teaser} in a given order. Trajectory generation under collision avoidance constraints is a typical application of the problem, where the polytopes represent the free configuration space of an UAV~\cite{gao18} or a car. 
In this case the time-parametrisation requires the computation of the time that the trajectory will spend in each of the polytopes.

\begin{figure}
\centering
  \begin{overpic}[width=0.99\linewidth]{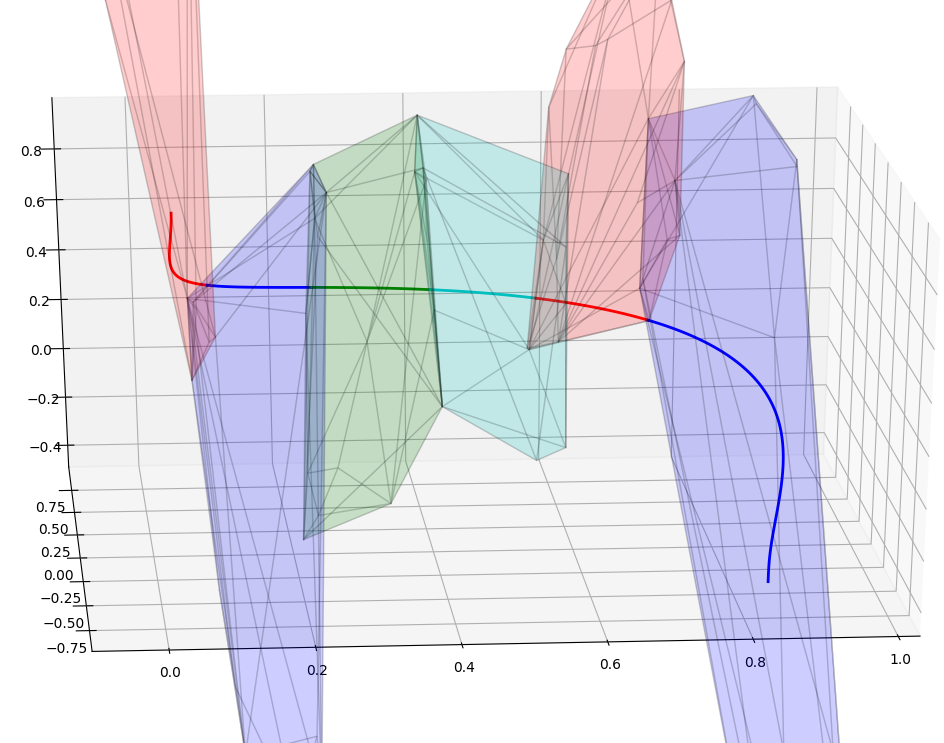}
	\end{overpic}
\caption{Illustration of the Polytope Traversal problem in 3D. The coloured polytopes correspond to constraints that concern specific parts of the Bezier curve displayed. The resulting curve is constrained on the initial / terminal positions and subject to constraints on the derivatives.}
		   \label{fig:teaser}
\end{figure}

\subsection{Current approaches for PT require an initial guess}
Under the (common) assumption that the dynamic constraints  are linear, if the time allocations are fixed the problem is convex. Likewise if the path is fixed, the problem is again convex\cite{Verscheure09}. This fact motivates the use of a decoupled approach to first compute a geometrically valid path, then the optimal time parametrisation of the path \cite{TOPP14, hauser2014fast}. While the decoupling implies that there is no guarantee of obtaining a global minimum, the reliability of this approach has been demonstrated.
Bilevel formulations that exploit the gradients of each of the problems can successfully be used to iteratively improve the results in spite of their non-linear structure\cite{Sun20, Tang20}.

Such approaches will be efficient under the assumption that the geometric path is always feasible, which is the case in several instances of the problem (including UAVs): 
if the dynamics of the system allow to accelerate freely in any direction and the problem has a solution, any geometric path will be feasible given arbitrarily large time 
allocations.

Unfortunately this assumption does not hold when the constraints make it impossible to accelerate in one direction. This happens in CROC~\cite{Fernbach:ccroc}, an instance of the problem where the trajectory represents the motion of the centre of mass of a legged robot, subject to switching contact constraints.
In such cases, providing a good initial guess for the bi-level optimisation can prove challenging. 

We are thus primarily concerned with the efficient computation of a feasible trajectory.
A feasible trajectory would provide a good initial guess to the aforementioned methods and could allow to include time as a variable for mixed-integer problems that try to compute the optimal polytope traversal order\cite{deits2015efficient}.
Furthermore, the efficient computation of a good initial guess is critical for the performances of sampling-based rejection methods that are only interested in the feasibility rather than the optimality to solve more complex problems~\cite{tsounis2020deepgait, tonneau-TRO18}.

\subsection{Contribution}

We observe that when the proportion of the trajectory spent in each polytope is specified, the PT problem can be formulated as a convex Quadratically Constrained Quadratic Program (QCQP). This allows
the simultaneous computation of the path and the total time of the trajectory. This formulation can be obtained by representing the trajectory as a Bezier curve of arbitrary degree and exploiting its De Casteljau decomposition. The formulation guarantees a minimum time trajectory for the given distribution.

To evaluate the success rate of our framework, we semi-randomly generate PT problems.
To identify problems that admit a solution, we implement an evolutionary strategy that efficiently samples the proportion of the trajectory spent in each polytope.
The CMA-ES algorithm that we use does not scale well as the number of polytopes grows but provides satisfying performances for problems with less than 5 polytopes, which are those 
that we target. 

We propose two contributions that advance the state of the art with respect to this objective:
\begin{itemize}
\item A convex formulation of the Polytope Traversal (PT) problem that simultaneously computes a path and its time parametrisation for the polytope traversal problem. 
The formulation is conservative but guaranteed to converge to locally optimal feasible solutions.
\item An evolutionary algorithm that exploits our convex formulation to find a better minimum time trajectory, in addition to determining the feasibility of a problem.
\end{itemize}

Our formulation and implementation are efficient (less than 100 ms are required to solve problems with less than 10 polytopes), and work in arbitrary dimension with polynomials of arbitrary degree. Our code is implemented using the NDCurves library \cite{ndcurve} and is entirely open source.

In the remainder of this paper, we first recall important notions on Bezier curves and provide additional definitions \sref{sec:def}. We then formalise the PT problem (Sections \ref{sec:polytope} and \ref{sec:problem}), before deriving a convex formulation of it \sref{sec:convex} and presenting our evolutionary strategy \sref{sec:cma}.
After presenting our experiments \sref{sec:impl} we discuss the results obtained \sref{sec:discussion}.

\section{Preliminaries}
\label{sec:def}

\subsection{A reminder on Bezier curves}
We first recall relevant properties of Bezier curves~\cite{de1959courbes}. 
\subsubsection*{Trajectory as a Bezier curve}
We define a trajectory   $\xv(t) , t \in [0,T]$ as a polynomial of arbitrary degree $n$ that takes its values in $\mathbb{R}^{dim}$, with $dim$ the dimension of the problem, $T \in \mathbb{R}^+$ the duration of the trajectory and $t$ a time parameter. 
Any polynomial can be written as a Bezier curve of the same degree $n$:\begin{equation*}
\label{eq:contribution:bezier_generic}
\xv(t) =   \sum_{i=0}^n B_i^n(t / T) \vc{x}_i  
\end{equation*}
where  the $B_i^n$ are the Bernstein polynomials and the $\vc{x}_i$ are the $n+1$ control points of the curve.
We also define the vector $\xv = [\vc{x}_0, \dots, \vc{x}_n]$ that contains all the control points.
The main variables of our problem will be the control points $\xv$ and the total time $T$.

The control points of any derivative of a Bezier curve are expressed as a linear combination of its control points.
We note their expression for the velocity $\dex(t)$ and acceleration $\ddx(t)$ curves of concern to us:
\begin{flalign*} 
\dex(t) =   \sum_{i=0}^{n-1} B_i^{n-1}(t / T) \vc{D}^1_i\frac{\vc{x}}{T} \\
\ddx(t) =   \sum_{i=0}^{n-2} B_i^{n-2}(t / T) \vc{D}^2_i\frac{\vc{x}}{T^2} 
\end{flalign*} 

with the $\vc{D}^1_i$ and $\vc{D}^2_i$ constant matrices of appropriate size.\footnote{In this paper we often introduce similar matrices and vectors. For brevity we do not introduce specific variables to specify their size. The number of rows is always problem dependent while the number of columns is equal to the size of the variables.} \\

\subsubsection*{Curve decomposition with the De Casteljau algorithm}
$\forall t_c \in [0,T]$ there always exists a decomposition of a Bezier curve into two curves $\xv(t)^0$ and  $\xv(t)^r$ such that:

\begin{flalign*} 
\forall t \in [0, t_c], \xv^0(t) = \xv(t) \\ 
\forall t \in [t_c, T], \xv^r(t) = \xv(t) 
\end{flalign*}

The continuity between the curves is $\mathcal{C}^\infty$ and their degree is also $n$. The curves are given by the De Casteljau algorithm and their control points are, as for the derivatives,
obtained as a linear combination of control points of $\xv(t)$:

\begin{flalign*} 
\label{castel}
\forall i \in \{0, \dots, n\}, \xv^0_i = \vc{C}^0_i\xv \\ 
\forall i \in \{0, \dots, n\}, \xv^r_i = \vc{C}^r_i\xv 
\end{flalign*}

with $\vc{C}^{\{0,r\}}$ constant matrices of appropriate size.
Any sub-curve can be decomposed with the same guarantees.

\subsection{Linear constraint definitions}
We now detail how any constraint considered can be written in as a linear combination of the control points $\xv$.

\subsubsection*{Convexity properties of Bezier curves}
A Bezier curve is entirely contained in the convex hull of its control points: $\forall t \in [0,T], \xv(t) \in conv(\vc{x}_0, \dots, \vc{x}_n)$, where $conv$ denotes the convex hull operation. Therefore, a sufficient condition to verify any constraint of the form 
\begin{equation*}  \forall t, \vc{B} \xv(t) \oplus \vc{b} \end{equation*} with $\vc{B}$  and $\vc{b}$ constant matrix  and vector of appropriate size and $\oplus$ describing either an equality ($=$) or inequality ($\leq$) constraint, is to simply verify  \begin{equation*} \forall i \in \{0, \dots, n\},  \vc{B} \xv_i \oplus \vc{b}. \end{equation*}
By stacking appropriately $n+1$ times $\vc{B}$ and $\vc{b}$ into the matrix $\vc{K}$ and vector $\vc{k}$, we can write the equivalent constraint:

\begin{equation}\label{kin} \vc{K} \xv \oplus \vc{k} \end{equation}
Although the condition is not necessary it is commonly used due to its practical interest, the main advantage being that it guarantees continuously that the curve satisfies the constraints. The current alternative consists in discretising the curve and evaluating the constraint at those discrete points, as is commonly done in numerical optimisation.
Our formulation works with either approach. In the remainder of the paper we assume that the continuous formulation holds. \\

\subsubsection*{Derivative constraints}
The trajectory can be constrained with respect to velocity and / or acceleration.

The velocity can be linearly constrained by a set of linear equations $\mathcal{L} := \{\xv \in \mathbb{R}^{dim} | \vc{L} \xv \oplus \vc{l} \}$.
We express the constraints in terms of the control points of $\dex(t)$: 
\begin{equation*} \forall i \in \{0, \dots, n-1\},   \vc{L} \vc{D}^1_i \frac{\xv}{T} \oplus \vc{l} \end{equation*}

We stack all the constraints in matrix and vector $\vc{V}$  and $\vc{v}$ of appropriate size to write the velocity constraints in a single block  and multiply by $T$ on both sides to obtain:

\begin{equation} \vc{V} \xv \oplus \vc{v} T \label{vel}\end{equation}

We can proceed similarly for the acceleration constraints and obtain constraints of the form:

\begin{equation} \vc{A} \xv \oplus \vc{a} T^2 \label{acc} \end{equation}

\subsubsection*{Geometric constraints}
As shown in \fg{fig:teaser}, our trajectory is constrained by a sequence of $m+1$ polytopes $\mathcal{H}^j, j \in \{0, \dots, m\}$ defined as 
\begin{equation}\label{H}\mathcal{H}^j := \{\vc{y} \in \mathbb{R}^{dim} | \vc{H}^j \vc{y} \leq \vc{h}^j\} \end{equation}
 
with $\vc{H}^j$ and $\vc{h}^j$ constant matrix and vector.

\subsection*{The CMA-ES algorithm}
In this work we compare our method with the CMA-ES~\cite{hansen2006cma} algorithm, which is our best available approximation of a ground truth that can be obtained with a reasonable amount of time. CMA-ES is a derivative free evolutionary algorithm that aims at finding a solution to an optimisation problem by sampling values of its variables. A population of samples is evaluated based on the cost function of the problem. A new population is the generated based on a stochastic variation of the best samples and this iterative process is repeated until a termination criteria is met.

\section{Handling of the polytope traversal constraints}
\label{sec:polytope}
We make use of the De Casteljau algorithm to express the constraints requiring $\xv(t)$ to belong to the polytopes $\mathcal{H}^j$, based on the following conservative assumption.

We define $s_j, 0 < s_j < 1$ the proportion of the total time spent by the trajectory $\xv(t)$ in $\mathcal{H}^j$, such that $\sum_{j=0}^{m}s_j = 1$.
Given the total time $T$, the time spent in $\mathcal{H}^j$ thus amounts to $T *  s_j $.

We define $\vc{s}= [s_0, \dots, s_m]$ and $s_{-1} = 0$.

By recursively applying the De Casteljau algorithm on our curves, we can define $m+1$ curves $\xv^j, 0 \leq j \leq m$:
\begin{flalign*} 
\forall j \in \{0, \dots, m\}, t \in [s_{j-1}*T, s_{j}*T] \Rightarrow \xv^j(t) = \xv(t) 
\end{flalign*}

All the control points of $\xv^j(t)$ are expressed as linear combination of $\xv$ as guaranteed by the De Casteljau algorithm.

We can now easily constrain a curve $\xv^j(t)$ to its assigned polytope $\mathcal{H}^j$ by constraining each of its control point using equation \eqq{H}:

\begin{align*} 
\forall i \in \{0, \dots, n\},  \vc{H}^j \vc{C}^j_i \xv \leq \vc{h}^j   \end{align*}

Stacking all the constraints we obtain $m+1$ constraints of the form:
\begin{flalign*} 
\forall j \in \{0, \dots, m\}, \vc{G}^j \xv \leq \vc{g}^j   \end{flalign*}

In summary, our trajectory is given by a Bezier curve of arbitrary degree. Assuming that the proportion of the time spent in each curve is given, every constraint
of our problem has the form of \eqq{kin}, \eqq{vel} or \eqq{acc}.

\section{Problem definition}
\label{sec:problem}
The inputs of our Polytope Traversal problem are:
\begin{itemize}
\item the set of  $m+1$ polytopes $\mathcal{H}^j$;
\item the proportion variable  $\vc{s}$;
\item optionally, a set of initial and terminal constraints on $\xv(t)$. They can be constraints on the initial / end positions with the form of \eqq{kin}; 
\item optionally, a set of dynamics constraints in the form of either \eqq{vel} or \eqq{acc}, possibly constraining the initial and terminal velocities / acceleration.
\end{itemize}

The objective is to find a minimum-time feasible trajectory $\xv(t)$ satisfying all the constraints.
Because $\vc{s}$ is fixed, the approach is conservative.

The general form of our problem is given by

\begin{align}
\label{eqn:TO}
\mathbf{find} \quad & \vc{x}, T & \\ 
\mathbf{min} \quad & T & \\
\st \quad &\vc{K} \xv  \oplus \vc{k}    \\
    \quad &\vc{V} \xv  \oplus \vc{v}T \label{tov}  \\
    \quad &\vc{A} \xv  \oplus \vc{a}T^2 \label{toa}
\end{align}
which is non-linear in the general case.

\section{Convexification of the PT problem}
\label{sec:convex}
We first note that for instances where constraint \eqq{toa} is missing, \eqq{eqn:TO} is a linearly constrained convex problem that can be solved with a Linear Program (LP) solver.
In the case where \eqq{tov}  is absent, we can simply replace variable $T^2$ with a scalar variable $y$ and minimize it:

\begin{equation*}
\begin{aligned}
\mathbf{find} \quad & \vc{x}, y & \\ 
\mathbf{min} \quad & y & \\
\st \quad &\vc{K} \xv  \oplus \vc{k}    \\
    \quad &\vc{A} \xv  \oplus \vc{a}y 
\end{aligned}
\end{equation*}

in which case $T = \sqrt{y}$.

Otherwise, we need an additional assumption to make the problem convex\footnote{One can also observe that if no equality constraints appear in the velocity constraints and that the velocity bounds include the null velocity, the formulation can remain linearly constrained and solved with a LP solver.}.

\subsection{An always feasible convex relaxation}
\label{sec:feasi}
We add the additional constraint $T \geq 1$, needed for our following proof. This can easily be satisfied without loss of generality by scaling $T$ appropriately.
We thus relax the problem \eqq{eqn:TO}:

\begin{align}
\label{eqn:TORelax}
\mathbf{find} \quad & \vc{x}, T, y  &  \\
\mathbf{min} \quad & y - T& \\
\st \quad &\vc{K} \xv  \oplus \vc{k}  \notag \\
    \quad &\vc{V} \xv  \oplus \vc{v}T \notag \\
    \quad &\vc{A} \xv  \oplus \vc{a}y \notag\\
    \quad &T^2  \leq y \label{relax} \\
    \quad &T \geq 1 \notag
\end{align}

Problem \eqq{eqn:TORelax} is a Quadratically Constrained Quadratic Program (QCQP)~\cite{Boyd:2004:CO:993483}, which is convex as \eqq{relax} is a quadratic convex constraint.
If the velocity constraints include equalities, then the solution of problem \eqq{eqn:TORelax} is valid for \eqq{eqn:TO} if and only if constraint \eqq{relax} is saturated, meaning that 
$T^2 = y$. Fortunately at the optimum this is always the case. We can also prove that the optimum of both problems is the same.

\textbf{Proof by contradiction:}
We consider $T_{opt} \geq 1$, the global minimum time trajectory for problem \eqq{eqn:TO} and the optimal control points $\xv_{opt}$. $\xv = \xv_{opt}$, $T = T_{opt}$ and $y = T_{opt}^2$ define a valid solution for problem \eqq{eqn:TORelax}. Indeed, the constraints on \eqq{eqn:TORelax} are a relaxation of the constraints on  \eqq{eqn:TO} with respect to $T$ and $\xv$, such that any solution of \eqq{eqn:TO} is a solution to \eqq{eqn:TORelax}.

Let $(T_{*}, y_*, \xv_{*})$ be an optimal solution for \eqq{eqn:TORelax}. 

We thus have three cases to consider:

\begin{itemize}
\item Either $T_{*} = T_{opt}$. Then it is clear that the minimum value $y - T_{*}$ is achieved when $y_* = T_{opt}^2$. Therefore the two problems share the same optimum;
\item Either $T_{*} < T_{opt}$. This means that the active constraint for $T$ in \eqq{eqn:TO} is  \eqq{toa}, implying that $y_* \ge T_{opt}^2 $. In such case $ T_{opt}^2 -  T_{opt} < y_* - T_{*}$, which contradicts the fact that $(T_{*}, y_*, \xv_{*})$ is optimal;
\item Either $T_{*} > T_{opt}$. As $y - T$ is bounded by the strictly increasing function $T^2 - T$ for $T \ge 1$, this means that $ y_* -  T_{*} > T_{opt}^2 -  T_{opt}$, which contradicts again the fact that  $(T_{*}, y_*, \xv_{*})$ is optimal\footnote{Note that we need the constraint $T \ge 1$ for this to be verified.}.
\end{itemize}

As a result, the only admissible case is the one where $T_{*} = T_{opt}$ and $y_{*} = T_{opt}^2$.$\square$

\section{Evolutionary strategy}
\label{sec:cma}
To test our approach we need a ground truth to determine the feasibility of the problems we consider. We implement the CMA-ES algorithm to sample values for the proportion variable $\vc{s}$, given as input to a slightly modified version \eqq{eqn:TORelax}, where we add a slack variable $\alpha \in \mathbb{R}^+$ to the inequalities:

\begin{equation}
\begin{aligned}
\label{CMA}
\mathbf{find} \quad & \vc{x}, T, y, \alpha  & \\ 
\mathbf{min} \quad & y - T + w * \alpha& \\
\st \quad &\vc{K} \xv  \oplus \vc{k}  \notag  \\
    \quad &\vc{V}^\leq \xv  \leq + \vc{1} \alpha + \vc{v}^\leq T \notag  \\
    \quad &\vc{A}^\leq \xv  \leq + \vc{1} \alpha + \vc{a}^\leq y \notag \\
    \quad &\vc{V}^= \xv  = \vc{v}^=T \notag  \\
    \quad &\vc{A}^= \xv  = \vc{a}^=y \notag \\
    \quad &T^2  \leq y  \\
    \quad &T \geq 1 \notag
\end{aligned}
\end{equation}


with $w$ a large positive scalar and the $\vc{V}^{=,\leq}$ terms denote the lines of $\vc{V}$ that are equalities (respectively inequalities). The resulting problem is always feasible if the original problem admits a solution and the cost conveniently feedbacks information regarding the extent to which the constraints are violated to the upper level. We empirically observed that CMA-ES converges faster with this formulation.

\section{Implementation and experiments}
\label{sec:impl}
Our code is entirely written in python, using the open source 
curve library NDcurves \cite{ndcurve}. NDcurves allows to compute automatically all the derivatives and the De Casteljau decomposition of the control variables.

The optimisation problems are solved using Gurobi \cite{gurobi}, while the CMA-ES algorithm is solved using the pycma library \cite{hansen2019pycma}. Although we report favourable computation times our primary objective is to measure the success rate of our approach in terms of feasibility.

\subsection{Problem generation}
We pseudo-randomly generate PT problems in 2D and 3D for which we check the feasibility using the CMA-ES algorithm.
A problem consists in a sequence of $m+1$ polytopes where each polytope intersects the one after it \fg{fig:teaser}, as well as randomly sampled initial and terminal states along with randomly sampled velocity inequality constraints.  The generation of the acceleration constraints is biased such that there is a probability $p = 0.4$ that deceleration along one or more axis is impossible. Such constraints are relevant for applications to legged locomotion \footnote{For instance, when the center of mass of the robot projection on the ground is outside of the support region defined by the effectors in contact it becomes impossible to accelerate towards the support region.}.

To determine a polytope we uniformly sample 10 points, compute their convex hull and translate the resulting polytope positively along the x direction and randomly along the others.
The translation amount along x is equal to a random number multiplied by the index $j$ of the polytope. To ensure that the polytopes share an intersection a random point is sampled along a line segment of two randomly generated points in two consecutive polytopes, and added to both polytopes~\fg{fig:teaser}.

The generation of the problem is obviously not a random process (which would be too inefficient). As a result the numbers presented here are not representative of all the instances of the PT problem. They are however shading a light on the potential benefits of the method.

\subsection{Testing variables}
In our tests we vary the number $m+1$ of polytopes that must be traversed, as well as the degree of the Bezier curve $\xv(t)$. The number of polytopes vary from 2 to 20, the degree from 2 to 20.

\subsection{Evaluation}

For each selected pair (number of polytopes, degree of the trajectory curve $\xv(t)$), we compute 100 feasible problems with the CMA-ES algorithm.
Each feasible problem is tested against problem \eqq{eqn:TORelax}. The ratio between the two numbers determine the success rate of the formulation. 

We use two means of initialisation for the proportion variable $\vc{s}$. One option is to heuristically define $\vc{s}$ by computing the shortest geometric path (ignoring the derivative constraints) and using the ratio of distance of each segment associated with one polytope over the total distance as the allocated proportion for each polytope, as commonly done \cite{gao18}. The other option simply consists in an even distribution of the proportion spent in each polytope.

We also compute the success rate of a naive initialisation that allocates arbitrary large times to each polytope.

\subsubsection{Success rate interpretation}

\begin{figure}[!b]
\centering
  \begin{overpic}[width=0.8\linewidth]{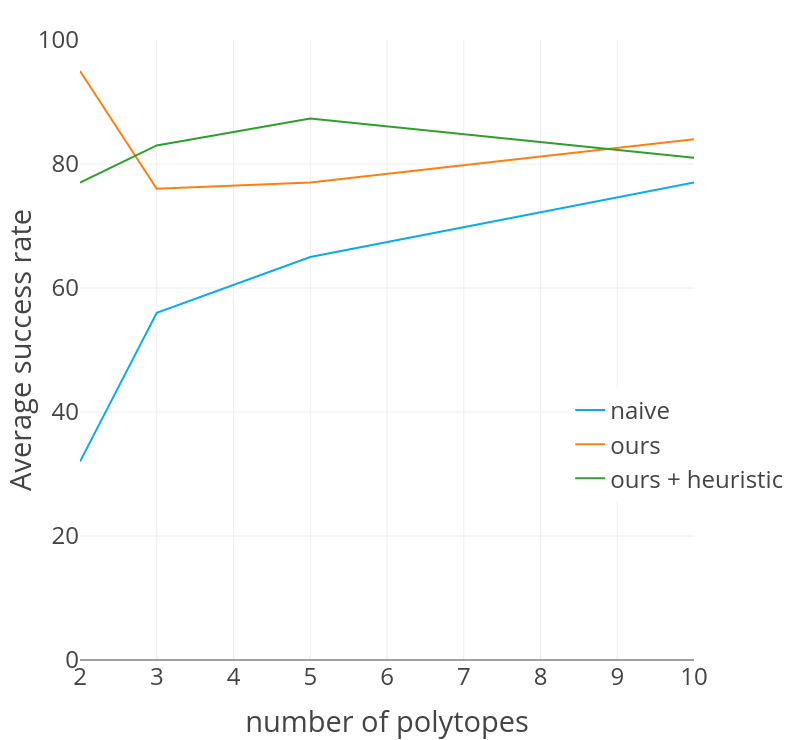}
 \end{overpic}
\caption{Averaged success rate of the different approaches as a function of the number of polytopes.}
     \label{fig:plot}
\end{figure}

%


\begin{table}[!b]
\centering

\begin{tabular}{|  P{3mm} | P{2mm} | P{2mm} | P{2mm} | P{2mm} | P{2mm} | P{2mm} | P{2mm} | P{2mm} | P{2mm} | P{2mm} | P{2mm} | P{2mm} | }
\cline{2-13}
\multicolumn{1}{c|}{}& \multicolumn{12}{c|}{Number of polytopes} \\
\multicolumn{1}{c|}{}& \multicolumn{3}{c|}{2} & \multicolumn{3}{c|}{3} & \multicolumn{3}{c|}{5} & \multicolumn{3}{c|}{10}\\ 
\hline
 Deg &{N}  &\tbf{O}& {Oh} & {N}   & {O}   & \tbf{Oh} &{N}  & {O}   & \tbf{Oh}  &{N}  & {O}   & \tbf{Oh}   \\
\hline
  5  & 31  & 96    & 80   & 56    & 66    & 80   & 64  & 71    & 90    & 59   & 78   & 64 \\
  10 & 31  & 94    & 77   & 62    & 80    & 85   & 63  & 80    & 85    & 75   & 83   & 88 \\
  20 & 34  & 93    & 74   & 51    & 84    & 85   & 68  & 81    & 87    & 75   & 92   & 92 \\
\hline
\end{tabular}\caption{Success rates against the CMA-ES ground truth given the degree of the trajectory and the number of polytopes to traverse. O and Oh refer to our approach initialised with or without the heuristic, while N refers to the Naive time allocation. Deg is the degree of the trajectory.}
\label{tab:succ}
\end{table}
The success rates are reported in Table~\ref{tab:succ}. We also present the success rate for each polytope, obtained by averaging the success rate according to the degree of $\xv(t)$ in \fgg{fig:plot}. 
We observe that our approach always outperforms the naive time allocation, but that the difference depends on the configuration of the scenarios.

For scenarios involving 2 or 3 of polytopes our approach is successful more than $80\%$ of the time in this context. These scenarios are those of particular interest for the CROC problem~\cite{Fernbach:ccroc} and suggest  that the method is particularly suited for such problems.

As the number of polytopes increases we observe that the minimum distance heuristic becomes almost immediately relevant. Our approach performs better with the heuristic for any number of polytopes above 2, while the success rate of the naive heuristic increases as well.

Likewise for small scenarios the degree of the curve has a limited influence on the success rate, suggesting that $\vc{s}$ is primarily responsible for determining the success of our approach. The degree of the curve plays a more determinant role as the number of polytopes increases.

As a conclusion, it appears that our approach is mostly relevant for scenarios involving a low number of polytopes, while providing for marginally superior results over the naive heuristic larger problems.

\subsubsection{Computation times}
We report the computation time required to solve \eqq{eqn:TORelax}, the naive Linear Program with fixed time allocation and CMA-ES in \fgg{fig:comp}.
For CMA-ES we report the computation time require to solve feasibility rather than the time to converge as the scales are not compatible. For the optimal problem, CMA-ES requires in average 3 seconds to converge for the smallest problems, one minute for handling problems with 10 polytopes and several minutes to handle scenarios with 20 polytopes.

Because our implementation is in Python the results  are useful for comparing the approaches but leave significant room for improvement. Computation times inferior to 10 ms were counted at 10 ms. As expected, CMA-ES scales rapidly with the number of considered variables, but provides decent performances for relatively small problems to solve for feasibility.
The naive approach and \eqq{eqn:TORelax} are similar problems in size, but the quadratic constraint makes the resolution of our approach slower, although the computation
times remain comparable.

\begin{figure}[!b]
\centering
  \begin{overpic}[width=0.8\linewidth]{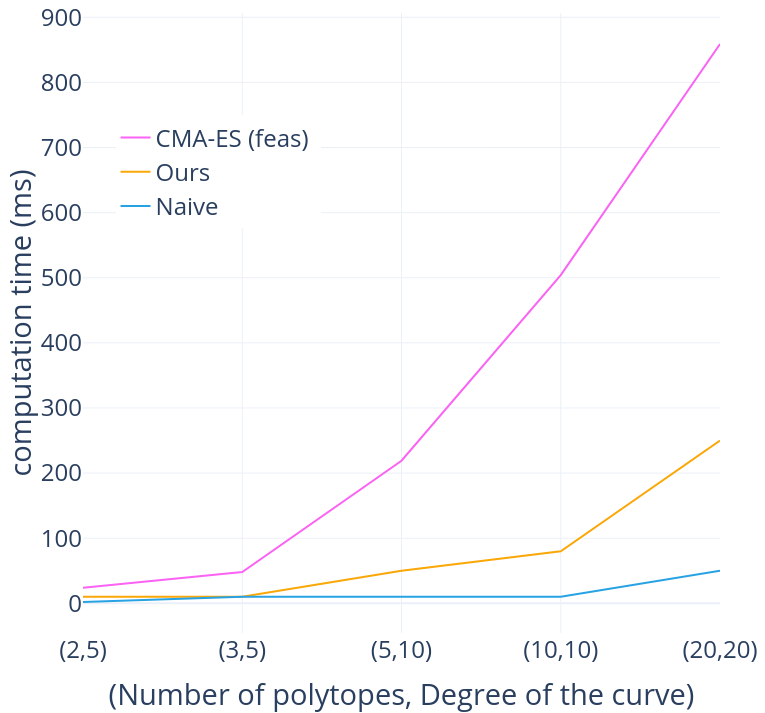}
 \end{overpic}
\caption{Computational performance of the methods averaged over 20 runs. The times given for CMA-ES are those required to find a feasible solution}
     \label{fig:comp}
\end{figure}

\subsubsection{Optimality}
The naive approach involves excessive traversal times by definition, thus it is only relevant to compare the times found by \eqq{eqn:TORelax} and CMA-ES.  We computed the average trajectory time to travel over scenarios comprising 2 and 5 polytopes and reported the results in Table~\ref{tab:optim}. As expected, our approach does not provide optimal times but the values are in the same order of magnitude as the approximated optimum. Again, these results do not have statistical significance and are illustrative. Solving for feasibility remains our objective.
\begin{table}[!b]
\centering

\begin{tabular}{  P{4mm} |  P{2mm} | P{2mm} | P{2mm} |}
\cline{2-4}
\multicolumn{1}{c|}{}&  \multicolumn{3}{c|}{Method} \\

\cline{2-4}
  &{C}  &O& {Oh} \\
\hline
\multicolumn{1}{c|}{Avg Min. time}& 24  & 33 & 28 \\ 
\hline

\end{tabular}\caption{Average minimum time obtained with CMA-ES (C) and our method with (Oh) or without the heuristic (O).}
\label{tab:optim}
\end{table}

\section{Discussion}
\label{sec:discussion}
Our experiments demonstrate that there are scenarios where finding a feasible initial guess for the Polytope Traversal problem is not trivial.
Those cases primarily involve a small number of polytopes with non-symmetric constraints on the derivatives. Our approach appears as a promising candidate 
for rapidly computing a feasible solution to such scenarios. Further research is required to demonstrate the interest of the approach for scenarios involving more polytopes. Indeed, it is quite possible that the reason why the naive approach performs better with larger sets of polytopes is mainly the result of the methodology to generate the problems, which may generate easier scenarios.

A strong advantage of the approach is that a problem instance can be solved in a single optimisation call. This makes it compatible with mixed integer solvers such as~\cite{deits2015efficient,tordesillas2019faster}. A significant part of the combinatorics could be removed by delegating the time and trajectory optimisation to \eqq{eqn:TORelax}, as it is guaranteed to provide the optimum for a given proportion allocation $\vc{s}$.

The proposed method also allows to provide convex approaches for locomotion~\cite{Fernbach:ccroc} with a mean to handle time as a variable when planning for the motion of legged robots, without breaking the convexity, thus providing an exciting avenue of research.

\section{Conclusion}
In this paper we proposed a conservative formulation of the Polytope Traversal problem that simultaneously computes the duration of the trajectory and the path it follows.

The method is convex and was proven to always converge to a locally optimal feasible solution. We have experimentally established the interest of the approach for generating feasible solutions over naive time allocation strategies.

Future work will investigate the possibilities offered by the approach in the context of robotics legged locomotion.

 
